\documentclass[letterpaper, 10 pt, conference]{ieeeconf}  
\usepackage{amsmath,amsfonts}
\usepackage{algorithmic}
\usepackage{algorithm}
\usepackage{multirow}
\usepackage{array}
\usepackage[caption=false,font=normalsize,labelfont=sf,textfont=sf]{subfig}
\usepackage{textcomp}
\usepackage{stfloats}
\usepackage{url}
\usepackage{verbatim}
\usepackage{graphicx}
\usepackage{caption}
\usepackage[backend=biber, sorting=none, style=ieee]{biblatex}
\IEEEoverridecommandlockouts                              

\title{\LARGE \bf
Dexterous Three-Finger Gripper based on Offset Trimmed Helicoids (OTHs)
}

\author{Qinghua Guan, Hung Hon Cheng and Josie Hughes
\thanks{Qinghua Guan$^{*}$ (qinghua.guan@epfl.ch), Hung Hon Cheng (hung.cheng@epfl.ch), Josie Hughes (josie.hughes@epfl.ch) are with the CREATE Lab, School of Engineering STI, EPFL, Swiss
        }
}

\begin{document}

\maketitle
\thispagestyle{empty}
\pagestyle{empty}

\begin{abstract}
This study presents an innovative offset-trimmed helicoids (OTH) structure, featuring a tunable deformation center that emulates the flexibility of human fingers. This design significantly reduces the actuation force needed for larger elastic deformations, particularly when dealing with harder materials like thermoplastic polyurethane (TPU). The incorporation of two helically routed tendons within the finger enables both in-plane bending and lateral out-of-plane transitions, effectively expanding its workspace and allowing for variable curvature along its length. Compliance analysis indicates that the compliance at the fingertip can be fine-tuned by adjusting the mounting placement of the fingers. This customization enhances the gripper's adaptability to a diverse range of objects. By leveraging TPU's substantial elastic energy storage capacity, the gripper is capable of dynamically rotating objects at high speeds, achieving approximately 60° in just 15 milliseconds. 
The three-finger gripper, with its high dexterity across six degrees of freedom, has demonstrated the capability to successfully perform intricate tasks. One such example is the adept spinning of a rod within the gripper's grasp.

\end{abstract}

\section{INTRODUCTION}

For many manipulation applications including manufacturing, agriculture and food handling, robots frequently encounter dynamic and evolving task requirements~\cite{chitroda2023review,tai2016state,zhang2020state,fontanelli2020reconfigurable}. Consequently, robotic grippers with high adaptability and responsiveness are crucial to adapt to changing task requirements~\cite{seguna2001mechanical, zhu2024dual}. Whilst rigid grippers can offer high force and precision, soft grippers leverage the inherent compliance of their materials to enabling grasping of a wider range of scales and shapes of objects~\cite{hughes2016soft, shintake2018soft,guan2020status}. However, most soft grippers are limited to simple grasping tasks ~\cite{cui2024design}, and their capabilities can not extend to more dexterous tasks.  

One approach to performing more dexterous tasks is developing more anthropomorphic inspired hands or reconfigurable grippers~\cite{kolluru2000design,spiliotopoulos2018reconfigurable, wang2023variable}. 
Although effective, these typically require a significant increase in the control and actuation complexity and hence associated cost. Reconfigurable soft grippers with variable gripping patterns have been proposed to enable higher adaptability and more complex manipulation tasks~\cite{zhou2017soft,zhong2019soft,low2021sensorized}, although their reconfiguration is predominantly actuated by rigid mechanisms. While enveloping grippers and multi-modal grippers~\cite{zhang2022design,guan2023multifunctional,jain2023multimodal} exhibit excellent adaptability to object geometries and mechanical properties, they still lack more dexterous manipulation capabilities such as in-hand rotation and spinning. As such, there is a need to extend the adaptive capabilities of soft grippers which to enable manipulation of objects in more dexterous and dynamic ways~\cite{kang2023grasping}. 

Developing soft or compliant structures where the directional compliance can be tuned of varied offers one means of enabling more complex output behaviours. 
In our previous research, we introduced the compliant trimmed helicoids (TH) structure which enables the independent regulation of bending and axial stiffness for soft manipulators with symmetrical workspace~\cite{guan2023trimmed}. The structure enables the control over the compliance of the end-effector over the workspace of the robot. However, the structure and tendon routing is fully symetric and not suited for dexterous manipulation. In this study, we propose a offset-trimmed helicoids (OTH) structure with a tunable offset deformation center to achieve a tailored workspace biased towards the inner side of the gripper, akin to human fingers, as shown in Fig.~\ref{fig: OTH Design}. 
By introducing two helically routed tendons into the structure, the finger can achieve both in-plane bending and lateral out-of-plane transitions, further expanding its workspace with variable curvature along its length. Furthermore, compliance analysis reveals that the experienced compliance at the fingertip within the workspace can be tuned by adjusting the mounting placement of the fingers, enhancing the gripper’s adaptability to different objects. Due to the large elastic energy storage in TPU, as opposed to softer silicone materials, the gripper can also achieve dynamic object rotation at high speeds (approximately 60° in 15 ms).  Through optimization of the design we show how the three finger gripper, with six actuated degrees of freedom, can successfully performs gripping tasks but also more dexterous ones such as spinning objects in hand.

In the following section we first introduce the methods and design of the OTHs, before showing the Experimental Results. We finish with a Conclusion and Discussion.

\section{Methods}

In this section we first introduce the OTH design and its characteristics, and then analyse the kinematics and workspace of the structure.

\subsection{The Design of Offset Trimmed Helicoids (OTHs)}

\begin{figure*}[t]
\centering
\includegraphics[width=0.85\textwidth]{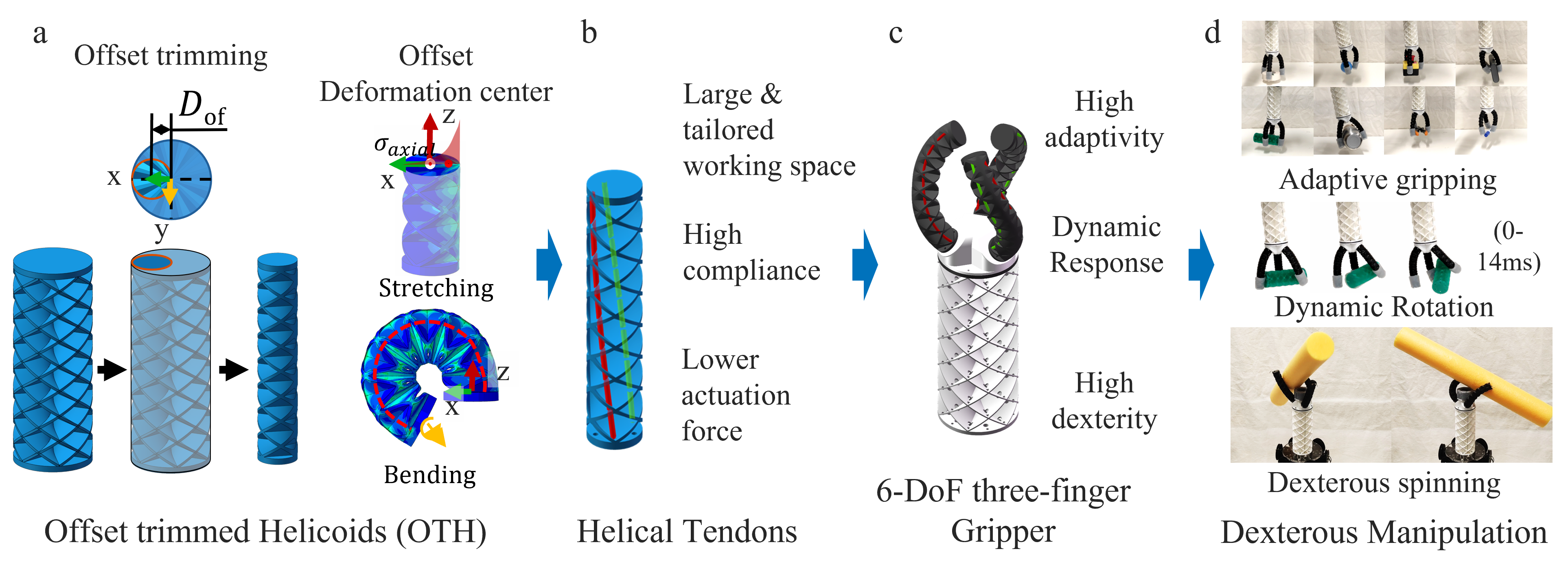}
\captionsetup{justification=centering}
\caption{The design of Offset Trimmed Helicoid (OTH) structure}
\label{fig: OTH Design}
\end{figure*}

\begin{figure*}[htb]
\centering
\includegraphics[width=0.85\textwidth]{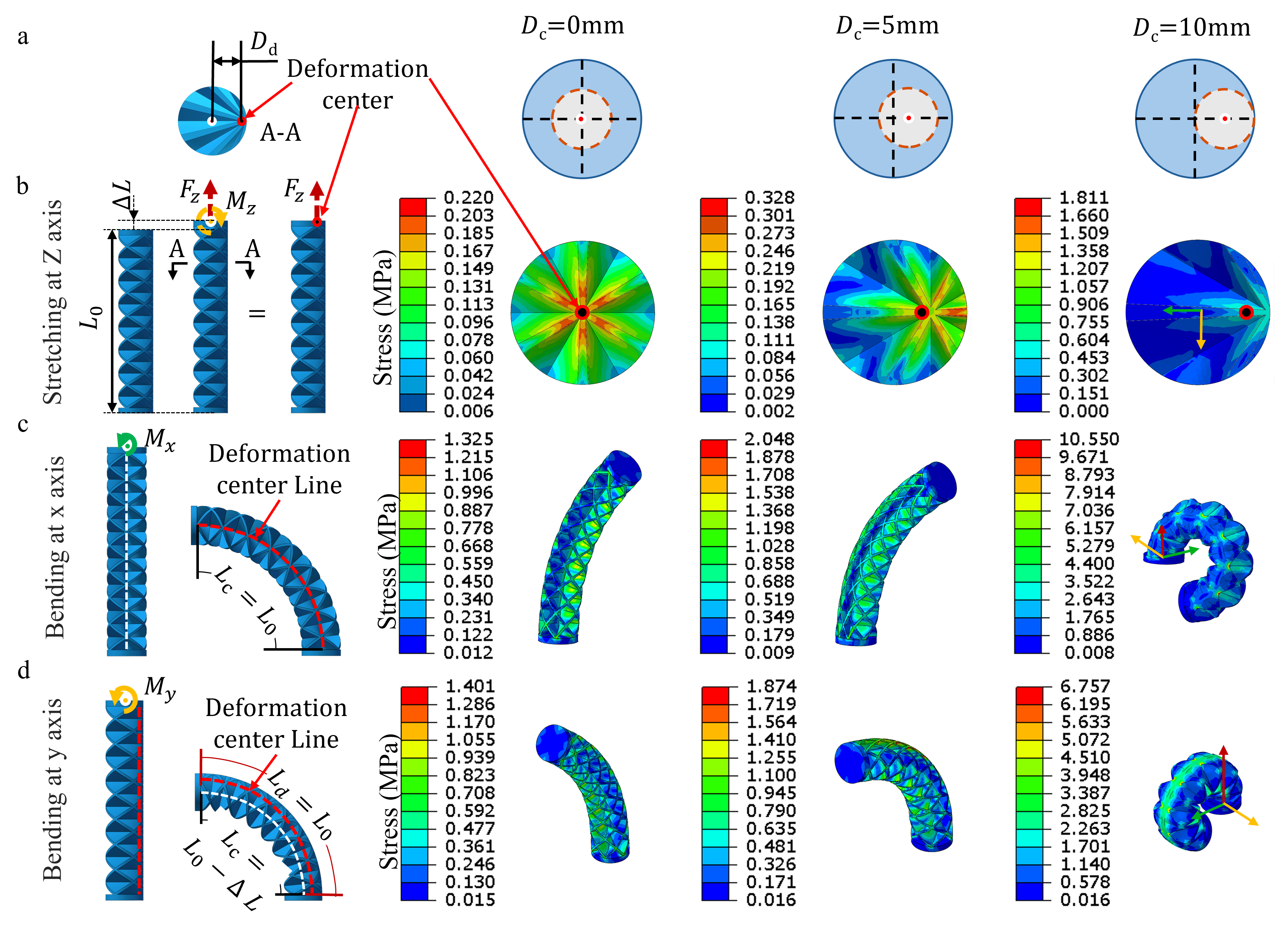}
\captionsetup{justification=centering}
\caption{FEM simulation of Offset Trimmed Helicoid (OTH) structure with different offset distances}
\label{fig: FEM simulation}
\end{figure*}

The OTH structure is generated by subracting a secondary cylinder offset by a distance $D_{c}$ from the center of a helicoid cylinder which comprises two groups of helicoids oriented in opposite directions, as depicted  in Fig. \ref{fig: OTH Design}a. To investigate the influence of $D_{c}$ on mechanical properties, finite element method (FEM) simulations were conducted using a linear elastic material with a modulus of 20 MPa and C3D10 elements approximately 1 mm in size, as illustrated in Fig. \ref{fig: FEM simulation}a-d. These FEM simulations were executed for OTH structures with identical trimming diameters, but with varying offsets.



As shown in Fig.\ref{fig: FEM simulation}, by varying $D_{c}$, the OTH structure can shift its deformation center from the section center. The deformation center is defined as the point on the cross-section where the reaction moment is zero during axial stretching or where the axial length remains unchanged under bending loads. Fig.\ref{fig: FEM simulation}b shows that for a non-uniform axial stress $\sigma_{a}$ along the $x$-axis generated with the axial load applied on the inhomogeneous structure, resulting in an axial force $F_z$ and a bending moment $M_y$ where their relationship can be expressed as
\begin{equation}
    \varepsilon_z = \frac{L_c}{L_0} ,  K_{ax,z} = \frac{F_z}{\varepsilon_z} , D_{d} = \frac{M_y}{F_z}
\end{equation}
where $\varepsilon_z$ is the nominal stretch strain and $ K_{a,z}$ is the axial stiffness in the $z$ axis. $D_{d}$ is the distance from the deformation center to the section centre.  Meanwhile, by applying bending moments $M_x$ and $M_y$ (Fig. \ref{fig: FEM simulation}c\&d) , the bending stiffness in $x$ and $y$ axis can be expressed as
\begin{equation}
    K_{b,x} = \frac{M_x}{k_x}, \quad K_{b,y} = \frac{M_y}{k_y}
\end{equation}
where $k_x$ and $k_y$ are the bending curvatures along the $x$ and $y$ axes and  respectively.
Due to the symmetry design in the $x$-$z$ plane, the bending in $x$ axis will not affect the center line length such that $L_c = L_0$ (Fig.\ref{fig: FEM simulation}c). However, the center line length is affected by the bending in $y$ axis such that $L_c = L_0 - \Delta L$ (Fig.\ref{fig: FEM simulation}d).

\begin{figure}[htb]
\centering
\includegraphics[width=3.5in]{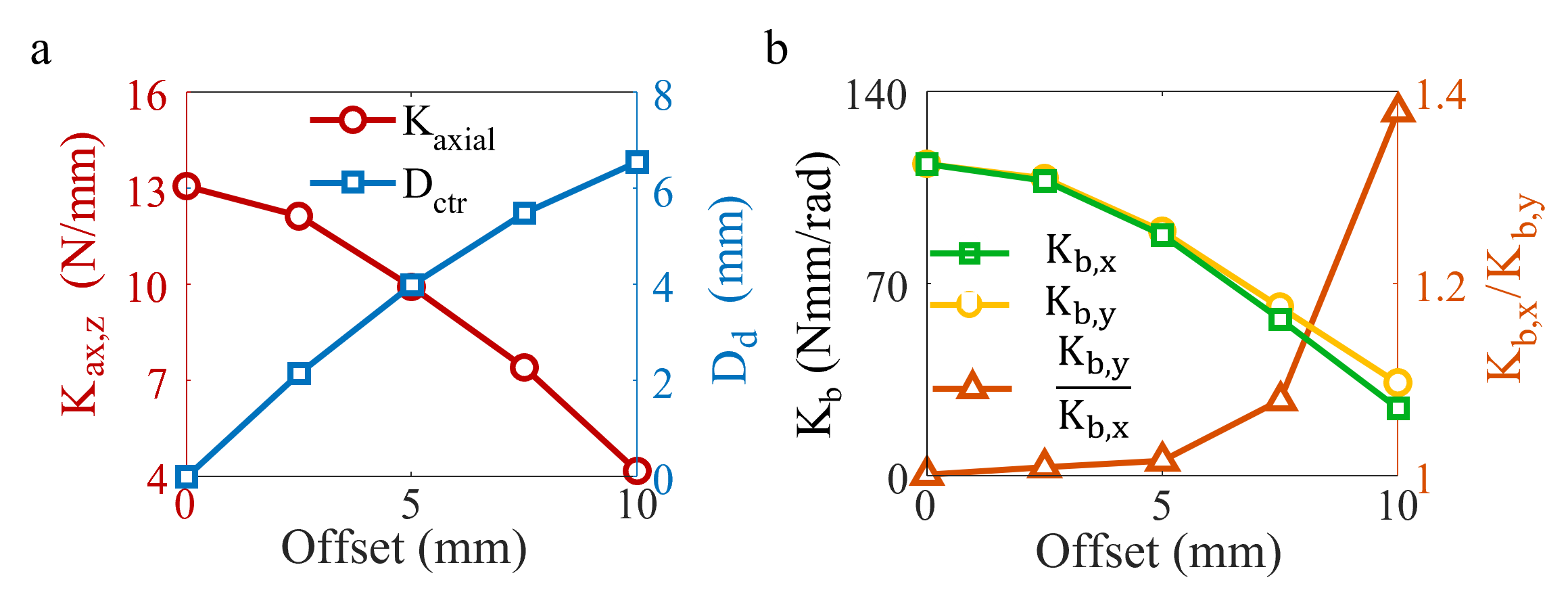} 
\caption{FEM simulation results of mechanical parameters}
\label{fig: FEM results}
\end{figure}

Then according to simulation results shown in Fig.~\ref{fig: FEM results}, it can be summarized as follows. 
  
\begin{itemize}
    \item $D_{c}=0$: Symmetrical structure, equal bending stiffness along $x$ and $y$ axes.
    \item $D_{c}\neq 0$: Asymmetrical structure, leading to unequal bending stiffness along $x$ and $y$ axes and a shifted deformation center. And as the offset distance $D_{c}$ increases, both axial stiffness and bending stiffness $K_{b,x}$ and $K_{b,y}$ decrease, while $D_{d}$ increases.
\end{itemize}
For instance, when $D_{c}$ reaches 10 mm, $D_{d}$ increases to 6.55 mm, and $K_{b,y}$ becomes 0.4 times greater than $K_{b,x}$ (Fig.~\ref{fig: FEM results}). Concurrently, the axial stiffness decreases to one-third of its original value, and the bending stiffnesses $K_{b,y}$ and $K_{b,x}$ decrease from 113 Nmm/rad to 33.8 Nmm/rad and 24.4 Nmm/rad, respectively, as shown in Fig,~\ref{fig: FEM results}.



\subsection{Kinematics of OTH}

\begin{figure}[htb]
\centering
\includegraphics[width=3.0in]{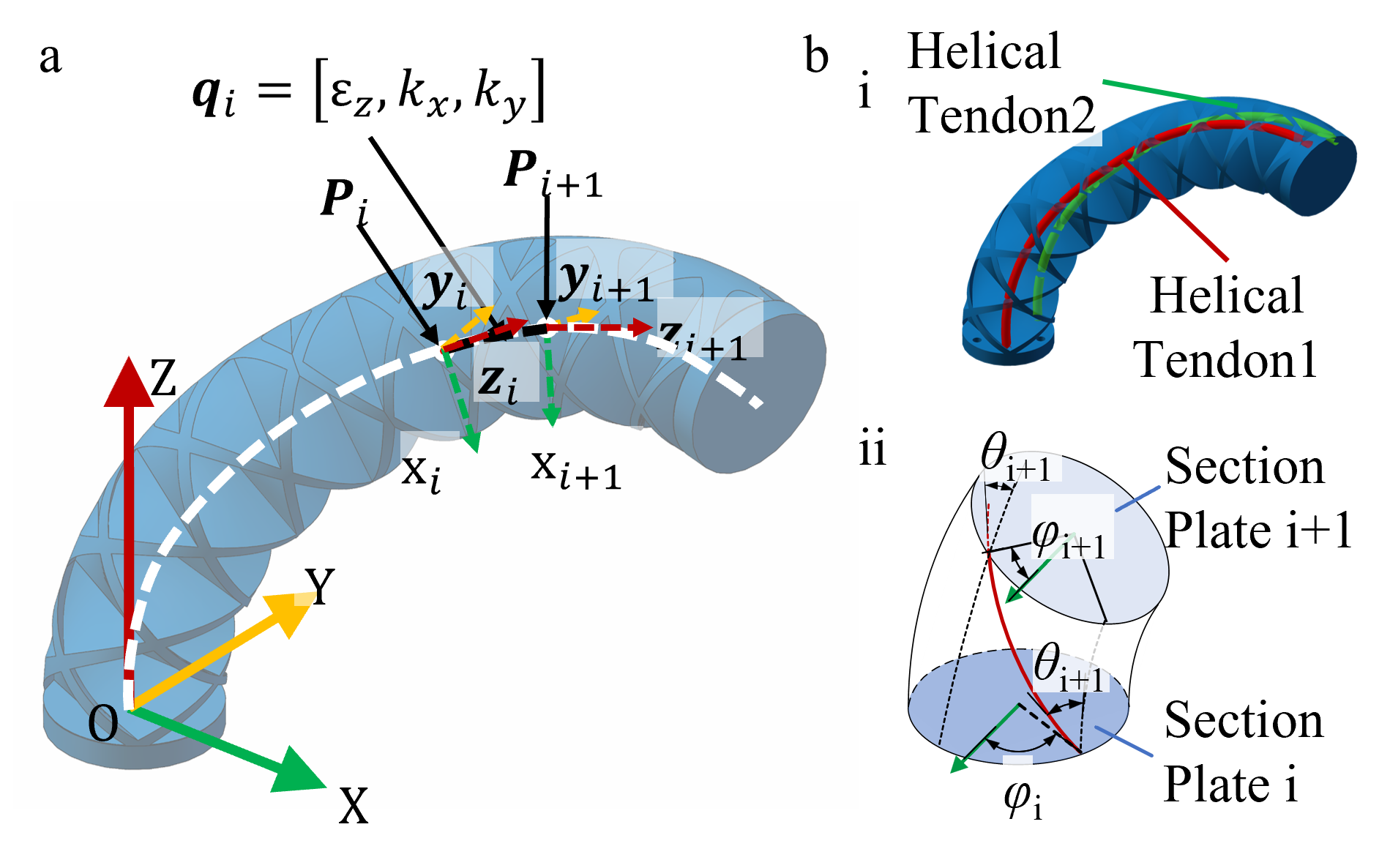}
\captionsetup{justification=centering}
\caption{Kinematics and actuation of OTH}
\label{fig: Kinematics and actuation of OTH}
\end{figure}

The Piecewise Constant Curvature (PCC) model is used to describe the kinematics  of the structure~\cite{webster2010design}.
The method divides the centre line of OTH into $n$ segment with constant strain and curvature, as shown in Fig.~\ref{fig: Kinematics and actuation of OTH}. Since the shearing and rotation deformation of OTH are negligible, the deformation at the $i^{th}$ segment can be described with the axial strain $\varepsilon_{z,i}$ and bending curvatures $k_{x,i}, k_{y,i}$ along $x$ and $y$ axis under the $\{o-xyz_i\}$  local frame at the section center $\mathbf{P}_i \in\mathbb{R}^{3}$.

The position and orientation of the OTH's tip, described in matrix form $\mathbf{S}_n \in \mathbb{R}^{4\times4}$, can be calculated by 
\begin{equation}  
\mathbf{S}_{n} = \prod_{i=0}^n\mathbf{T}_{i}
\end{equation}
where the transformation matrix $\mathbf{T}_{i} \in \mathbb{R}^{4\times 4}$ between two continue segment can be expressed as 
\begin{equation}
\label{eq: kinematics} 
\mathbf{T}_i =\left[\begin{matrix}\mathbf{R}_i, &\mathbf{P}_i - \mathbf{P}_{i-1} \\
0_{1\times 3} & 1\\ \end{matrix}\right]
\end{equation}
where $\mathbf{R}_i \in \mathbb{R}^{3\times 3}$ is the rotation matrix between two frames.  



\subsection{Actuation Force in Helicoid OTH Finger}
The soft robotic finger can be actuated embedded tendons arranged either in parallel to the central axis or helically (Fig. \ref{fig: Kinematics and actuation of OTH}b), allowing for variations in workspace dimensions and compliance characteristics. Compared to parallel tendons, helical tendons enable variable bending curvature and orientation along the axis, resulting in greater bending with reduced actuation force



The generic force acting on the $i^{th}$ segment of OTH can be calculated by the stiffness matrix $\mathbf{M}_{s} \in \mathbb{R}^{3\times 3}$ and the deformation $\mathbf{q}_i = [\varepsilon_{z,i},k_{x,i}, k_{y,i}]^T$ using the virtual work principle of:
\begin{equation}
    \label{eq: Mechanics}
\mathbf{F}_i={[F_z,M_x, M_y]}^T=\mathbf{M}_{s}\mathbf{q}_i \in \mathbb{R}^{3\times 1}
\end{equation}
where 
\begin{equation}
\mathbf{M}_{s} = 
\begin{bmatrix}
K_{a,z} & 0 & K_{a,z} D_{d} \\
0 & K_{b,x} & 0 \\
K_{a,z} D_{d} & 0 & K_{b,y} D_{d}^2
\end{bmatrix}
\end{equation} 
The relationship between the generic force $\mathbf{F}_i$ and the piecewise force $\mathbf{F}_{A,i}$ acting on the start point of the $i^{th}$ segment can be considered as $\mathbf{F}_i = (\mathbf{F}_{A,i}+\mathbf{F}_{A,i+1})/2$, where $\mathbf{F}_{A,i}$ can be expressed as
\begin{equation}
    \mathbf{F}_{A,i} = F_T \cos(\theta_i)[1,~sin(\phi)R_{T,i},~cos(\phi)R_{T,i}]^T
\end{equation}
where $\theta_i$ and $\phi_i$ represent the helical and azimuthal angles of the tendon at the $i^{th}$ section, and $R_{T,i}$ is the radial distance of the tendon from the section center.




\subsection{Reachable Workspace Analysis}

\begin{figure}[htp]
\centering
\includegraphics[width=0.75 \linewidth]{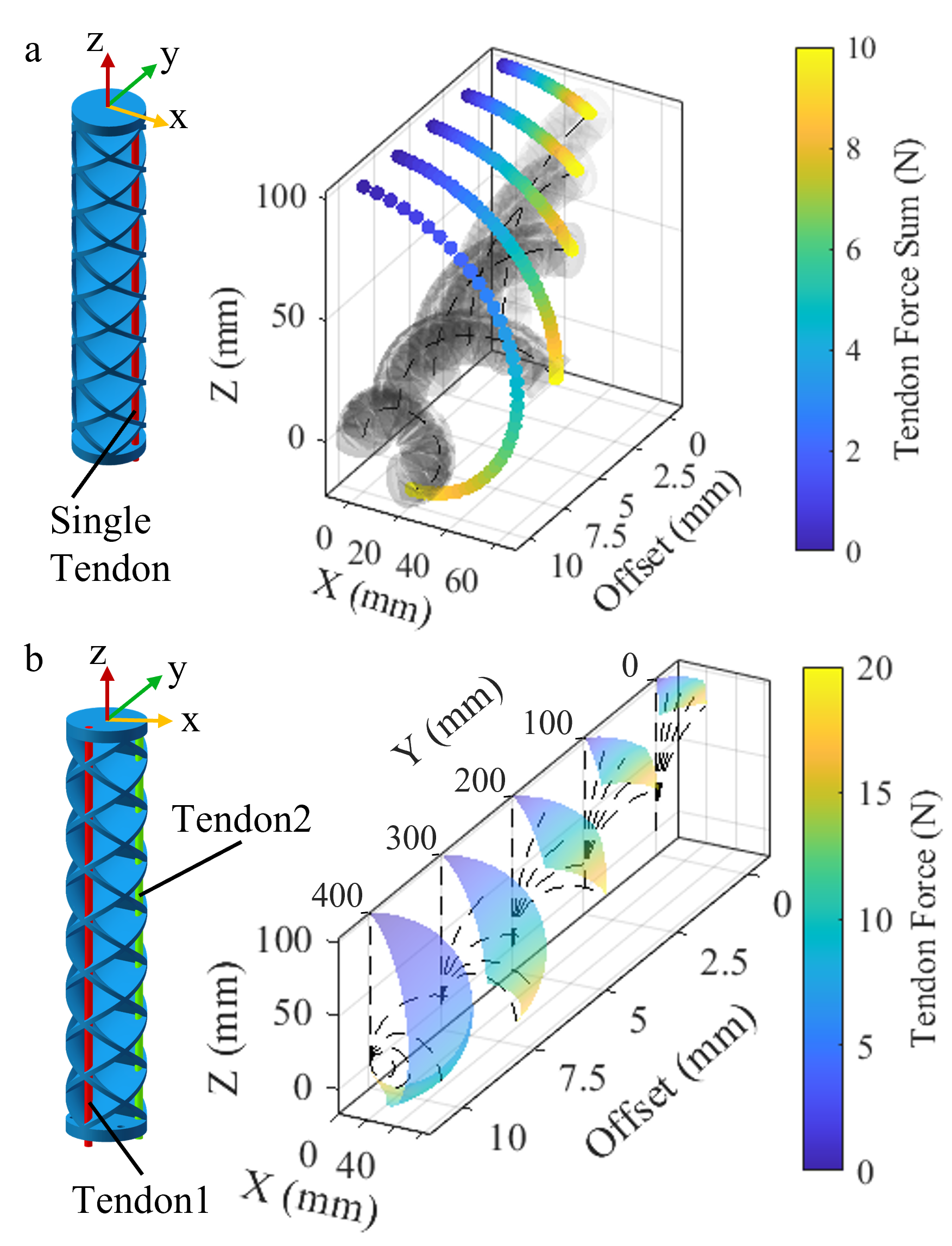}
\captionsetup{justification=centering}
\caption{Workspace of OTHs fingers with parallel tendons}
\label{fig: Parallel Tendon WSP}
\end{figure}

\begin{figure}[htp]
\centering
\includegraphics[width=0.85 \linewidth]{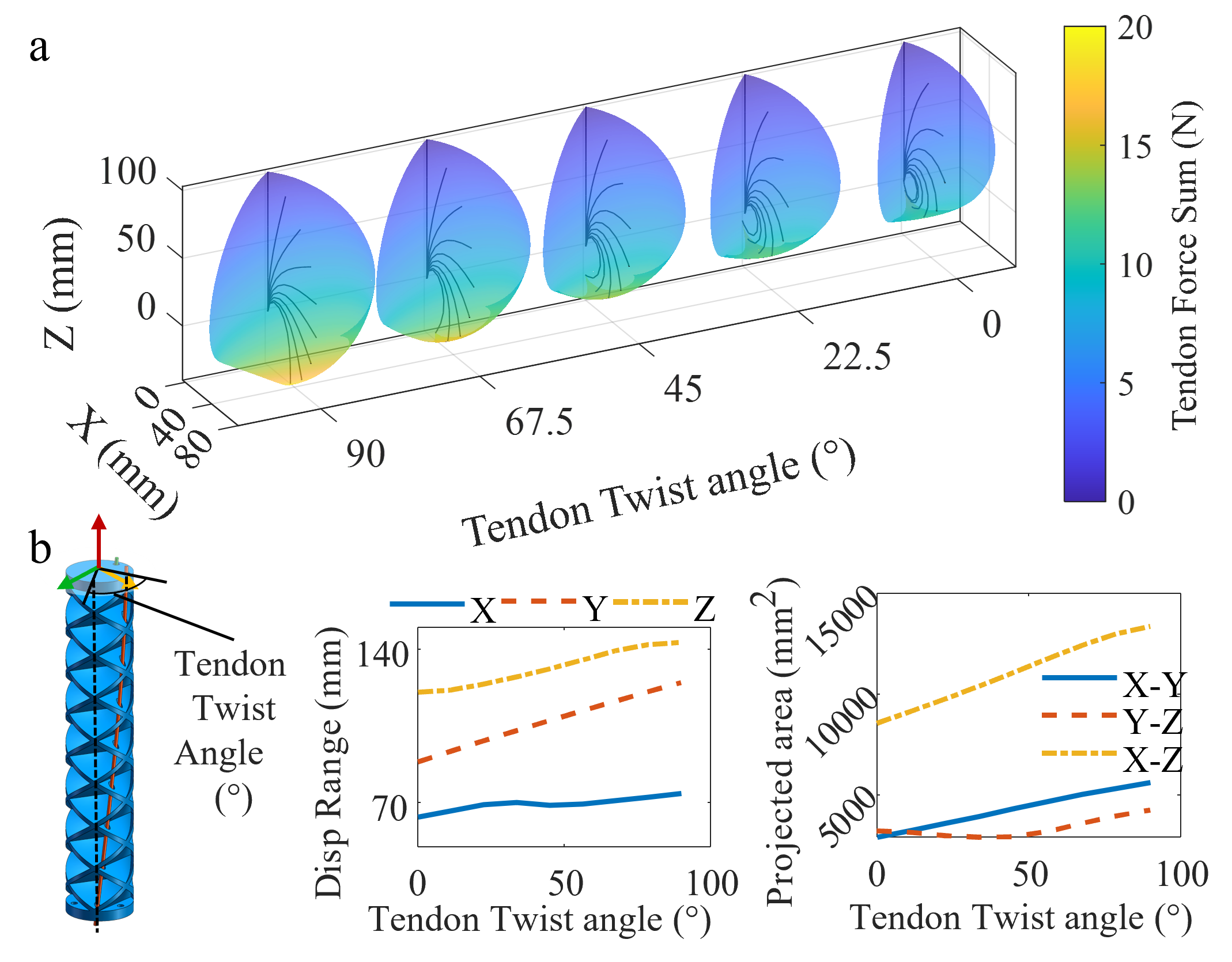}
\caption{Workspace of OTHs fingers with helical Tendons}
\label{fig: Helical Tendon WSP}
\end{figure}

To compare the characteristics of parallel and helical tendon configurations, tension $F_T$ from 0 to 10 N was applied to each tendon, with the tendons placed at $R_{T,i} = 8$ mm from the finger's axis at the base. As shown in Fig.\ref{fig: Parallel Tendon WSP}, Fig. \ref{fig: Helical Tendon WSP} and Table \ref{tab:my-table}, the maximum reachable workspace in the $x$-$y$-$z$ dimensions varies with different offset distances, tendon configurations, and tendon twist angles.

Initially, a single tendon was positioned parallel along the axis. As the offset distance increased from 0 to 10 mm, the z-axis displacement and bending range of the OTH finger expanded to 119.0 mm and 246°. Subsequently, the two tendons, placed on each side of the x-z plane, enabled the three-dimensional movement and expanded workspace in almost all directions. Finally, two tendons with opposite helical angles were deployed. The workspace expanded with increasing twisting angles and reached maximum displacement and projected area at a 90° twisting angle.

Based on the simulation results, the OTH finger with a 10 mm offset and a 90$^{\circ}$ twisting angle was selected for the gripper due to the larger reachable workspace.


\begin{table}[h!]
\resizebox{\columnwidth}{!}{%
\begin{tabular}{|c|c|c|ccc|}
\hline
\multirow{2}{*}{\begin{tabular}[c]{@{}c@{}}Tendon\\ Config.\end{tabular}} & \multirow{2}{*}{\begin{tabular}[c]{@{}c@{}}Motion\\ Area\end{tabular}}&\multirow{2}{*}{\begin{tabular}[c]{@{}c@{}}Tuning\\ parameter\end{tabular}} & \multicolumn{3}{c|}{\begin{tabular}[c]{@{}c@{}}Max Displacement \end{tabular}} \\ \cline{4-6} 
 &  & & \multicolumn{1}{c|}{x (mm)} & \multicolumn{1}{c|}{y (mm)} & z (mm) \\ \hline
\begin{tabular}[c]{@{}c@{}}One Tendon\\ (Parallel at \\$\phi = 90^\circ$)\end{tabular} & Planar& \multicolumn{1}{c|}{\multirow{2}{*}{\begin{tabular}[c]{@{}c@{}}Center offset \\ $D_c = 0$ \\to $10$(mm)\end{tabular}}} & \multicolumn{1}{c|}{\multirow{2}{*}{\begin{tabular}[c]{@{}c@{}}65.1\\ $(D_c = 7.5)$\end{tabular}}} & \multicolumn{1}{c|}{0} & \multicolumn{1}{c|}{\begin{tabular}[c|]{@{}c@{}}119.0\\ $(D_c = 10)$\end{tabular}} \\ \cline{1-2} \cline{5-6} 
\begin{tabular}[c]{@{}c@{}}Two Tendons\\ (Parallel at \\$\phi = \pm 60^\circ$ )\end{tabular} & Spatial &  & \multicolumn{1}{c|}{} & \multicolumn{1}{c|}{\begin{tabular}[c|]{@{}c@{}}88.8\\ $(D_c = 10)$\end{tabular}} &\multicolumn{1}{c|}{\begin{tabular}[c|]{@{}c@{}}119.0\\ $(D_c = 10)$\end{tabular}}  \\ \hline

\begin{tabular}[c]{@{}c@{}}Two Tendons\\(Helical,\\ starting at \\ $\phi = \pm 60^\circ$)\end{tabular} & Spatial&\begin{tabular}[c]{@{}c@{}}Twist angle\\ $\theta = 0$\\ to $90(^\circ)$\end{tabular}  & \multicolumn{1}{c|}{\begin{tabular}[c|]{@{}c@{}}74.4\\ $(\theta = 90^\circ)$\end{tabular}} & \multicolumn{1}{c|}{\begin{tabular}[c|]{@{}c@{}}124.7\\ $(\theta = 90^\circ)$\end{tabular}} & \multicolumn{1}{c|}{\begin{tabular}[c|]{@{}c@{}}143.0\\ $(\theta = 90^\circ)$\end{tabular}} \\ \hline
\end{tabular}%
}
\caption{Comparison between parallel and helical driven tendon with different offset $D_c$ and Twist angle $\theta$}
\label{tab:my-table}
\end{table}

\section{Experimental Setup}
\begin{figure}[htp]
\centering
\includegraphics[width=3in]{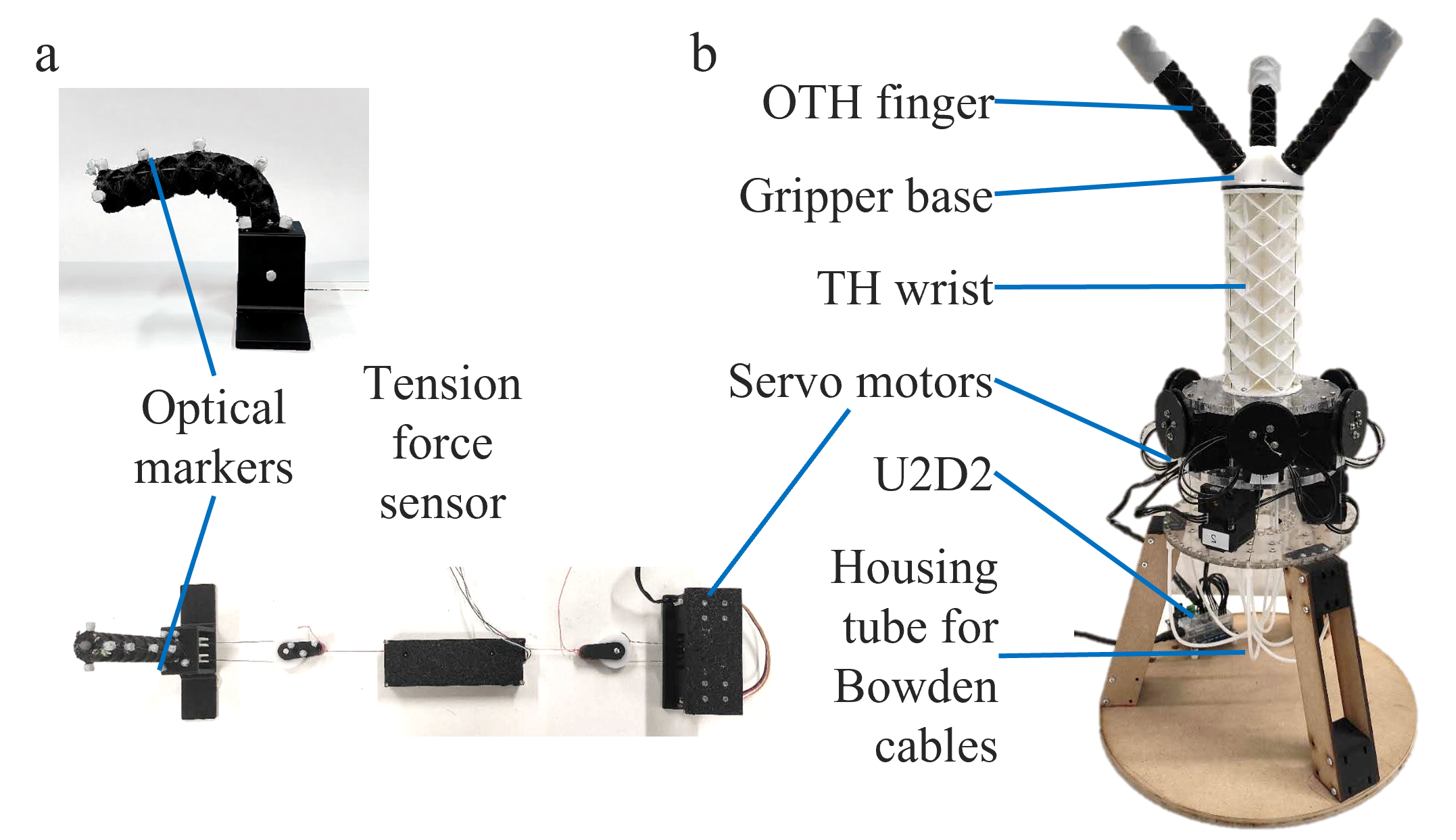}
\caption{Experimental Setup for finger and gripper tests}
\label{fig: Exp setup}
\end{figure}

In the experiment, three OTH structures each 3D-printed with TPU filament (98A, Purefil) and a length of 106 mm, were mounted on a base with a 45$^\circ$ tilt. Each finger was actuated by two position-controlled servo motors (Dynamixel XL430-W250, ROBOTIS INC), as shown in Figure \ref{fig: Exp setup}, forming a gripper. The workspace and actuation force were verified using a Motion Capture System (OptiTrack) and a tension force sensor.




The gripper was regulated utilizing the DYNAMIXEL U2D2 interface and its accompanying software development kit (SDK). The finger movements were classified into two categories: inner bending and lateral transition, to modify the gripper’s gestures. Hard-coded motion Primitives were devised for tasks such as grasping symmetric, asymmetric, and twisting objects, among others, to illustrate their functionality.

\section{Results}
\subsection{Finger performance}
\subsubsection{Finger characterization}
\begin{figure}[htb]
\centering
\includegraphics[width=1\linewidth]{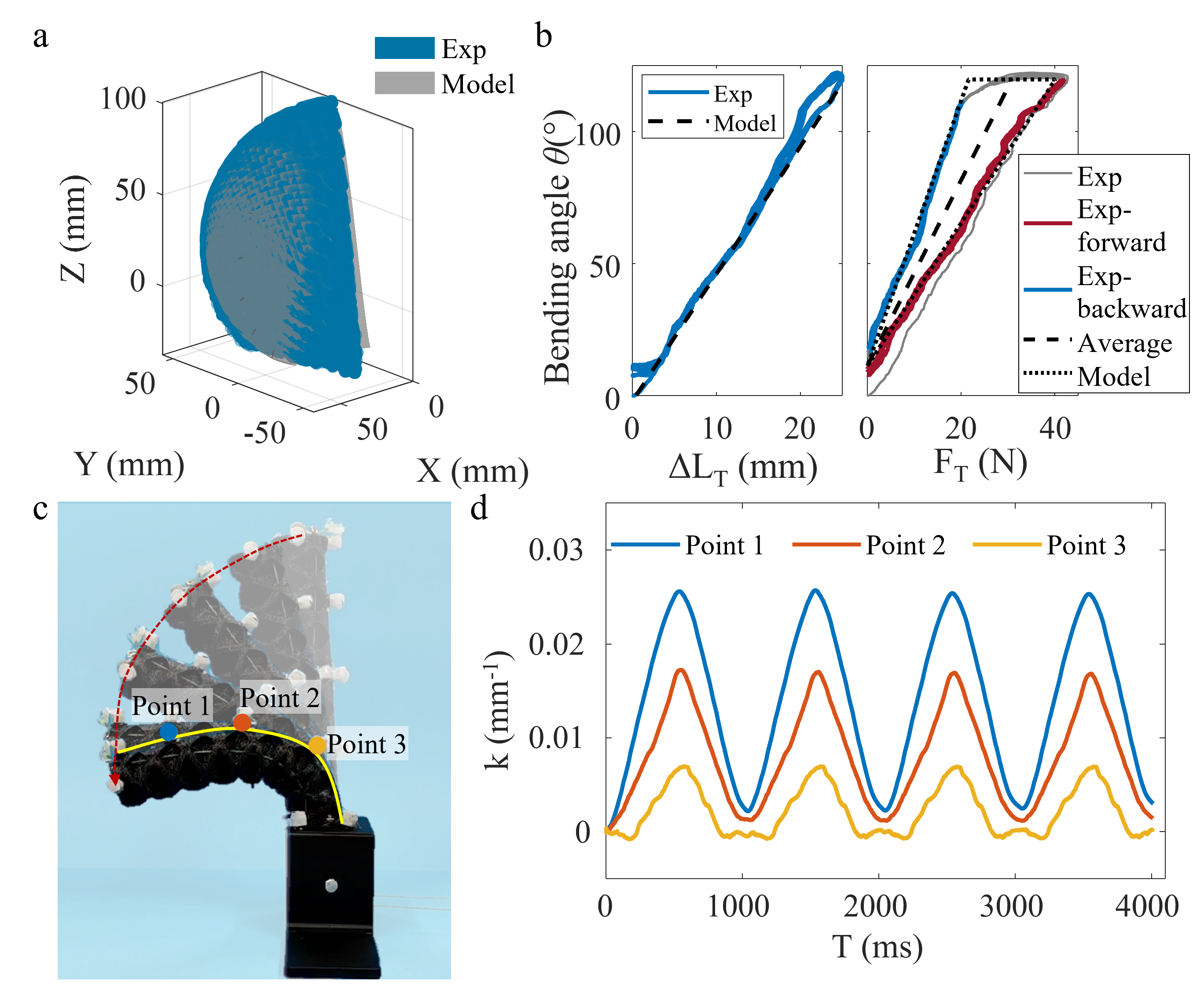}
\caption{OTH Finger Characterization}
\label{fig: Finger Characterization}
\end{figure}

The workspace of the OTH finger was characterized  and compared with the model results in Fig. \ref{fig: Finger Characterization}a\&c, showing good agreement. 

Subsequently, a cyclic bending test was performed, recording tendon contraction ($\Delta L_T$), tendon force ($F_T$), and finger deformation (Fig. \ref{fig: Exp setup}a and Fig. \ref{fig: Finger Characterization}b). Under the influence of friction and the material viscosity, the finger deformation exhibited a drift following cyclic loading (Fig. \ref{fig: Finger Characterization}b\&d). However, the bending angle demonstrated a significant linear relationship with tendon contraction, while a notable hysteresis effect was observed between the bending angle and tendon force, primarily due to tendon friction. A simple model could be derived for this relationship as follows:
\begin{equation}
\label{eq: Friction modeling}
F_T=(a\theta_b+b)(1+\mathrm{sgn}(\dot{\theta}_b)f_{fric}) 
\end{equation}
where $\theta_b$ is the bending angle, $a=0.2843$ and $b=-3.1852$ are the linear relationship factor between $\theta_b$ and the tension force $F_T$. The friction $f_{fric}$ defines the hysteresis effect due to tendon friction. Thus, based on the similarity principle, the nominal elastic modulus of the real OTH finger ($E_{real}$) can be derived as 64.85 MPa, according to the simulated OTH finger elastic modulus ($E_{sim}$) of 20 MPa and the linear relationship factor $a=0.0877$ for $F_T$ to $\theta_b$. 

Furthermore, Fig. \ref{fig: Finger Characterization}c\&d shows the variable curvature along the finger's length, which could lead to sequential bending during actuation, considering the blocking effect after the self-contact of the finger helicoids.

\subsubsection{Finger compliance}
Despite being actuated by two inextensible tendons, the OTH structure retained passive compliance due to its infinite degrees of deformation, allowing object-adaptation during gripping. This compliance is critical to performance, making its analysis within the workspace valuable for gripper design.

\begin{figure}[htb]
\centering
\includegraphics[width=\linewidth]{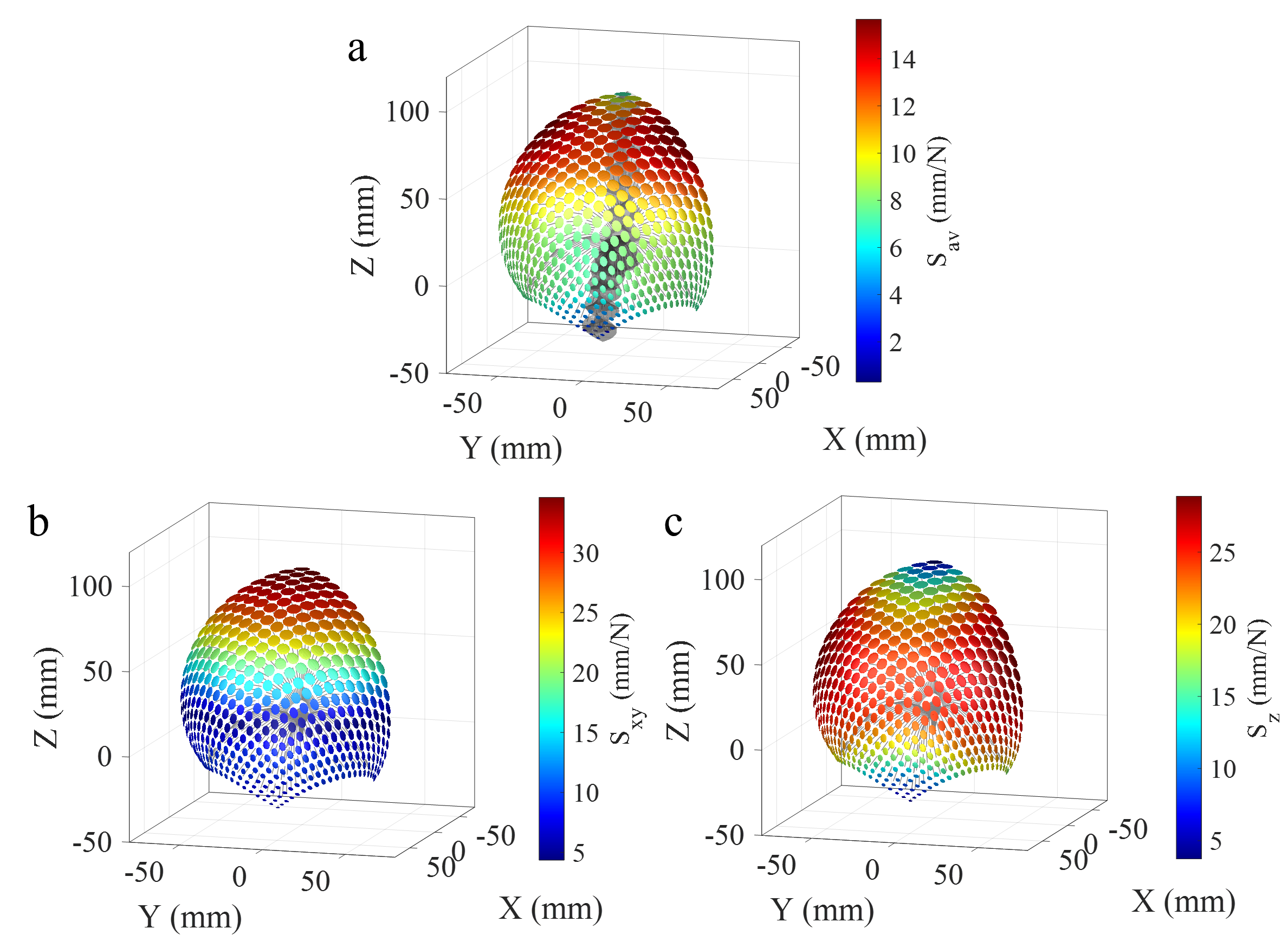}
\caption{Compliance distribution of OTH finger in different dimensions}
\label{fig: OTH Compliance}
\end{figure}

Passive deformation was simulated by applying a unit disturbance load ($\delta F_d$) in three orthogonal directions, with tendons held constant for each valid workspace configuration. The compliance in each state is then derived as follows:
\begin{equation}
\mathbf{S}_{c,i} = \mathbf{\delta D_d}/\mathbf{\delta F_d} \nonumber
\end{equation}
\begin{equation}
S_{c,av} = |\mathbf{\lambda}_{c,i}|, \quad S_{c,z} = |\mathbf{S}_{c,i}(:,3)|\nonumber
\end{equation}
\begin{equation}
S_{c,xy} = |\mathbf{S}_{c,i}(:,1) \times \mathbf{S}_{c,i}(:,2)| \nonumber
\end{equation}
where $\mathbf{\lambda}_{c,i}$ is a vector of eigenvalues $\lambda_{c,i,j}$ related to the eigenvectors $\mathbf{\nu}_{c,i,j=1,2,3}$ of the compliance matrix $\mathbf{S}_{c,i}$. $S_{c,av}$ is the averaged compliance at all directions. $S_{c,xy}$ is the averaged compliance at $x$-$y$ plane. $S_{c,xy}$ is the compliance at $z$ direction, as show in Fig. \ref{fig: OTH Compliance}.

\subsection{The three-finger gripper}
\begin{figure}[htb]
\centering
\includegraphics[width=\linewidth]{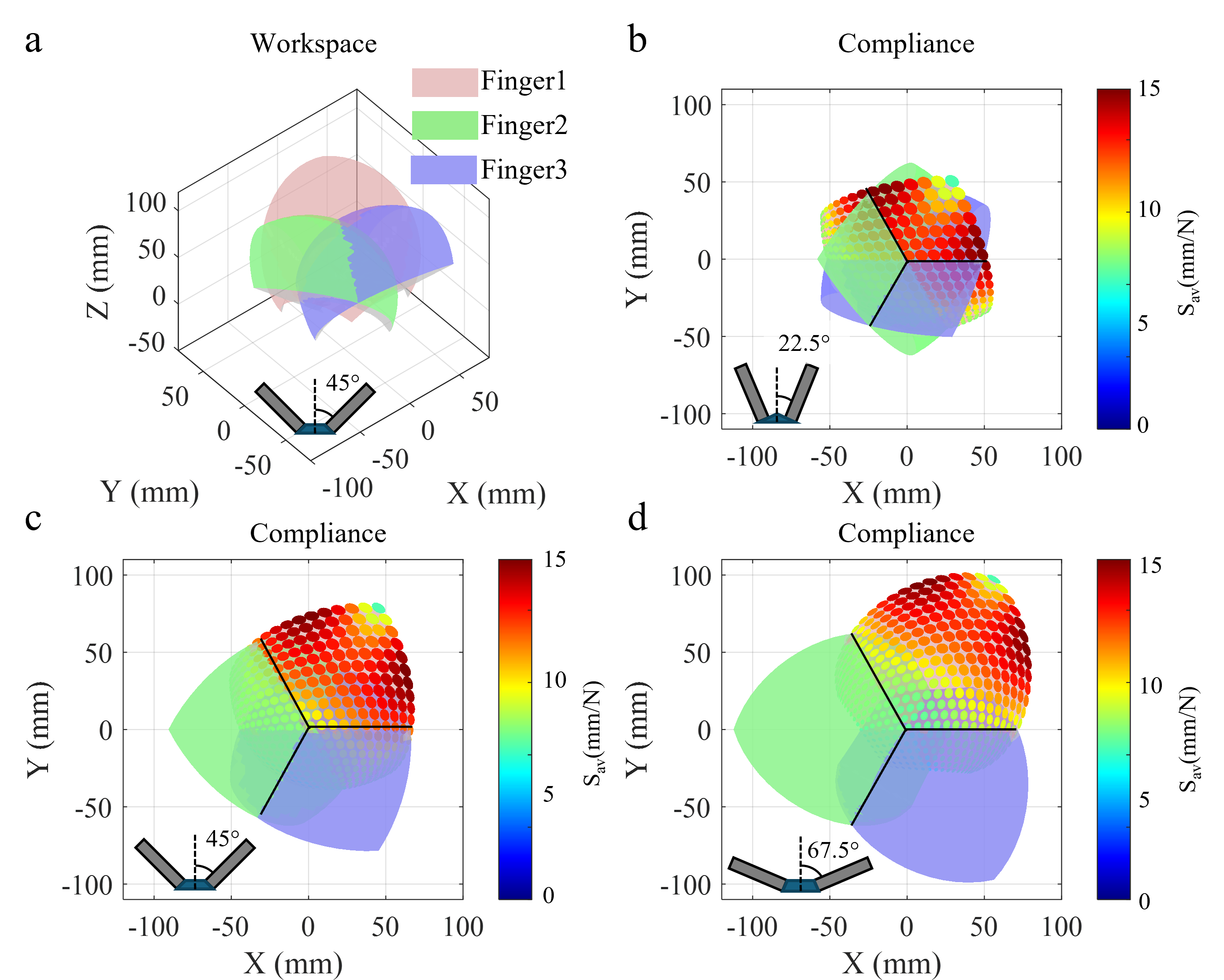}
\caption{The compliance distribution of grippers with different tilt angles}
\label{fig: three-figure compliance distribution}
\end{figure}

The workspace and compliance distribution were analyzed to optimize the tilt angle of the three fingers, as shown in Fig. \ref{fig: three-figure compliance distribution}. At a 22.5$^\circ$ tilt, the tip compliance within the workspace is high. However, significant overlap between the fingers reduces the overall workspace. At 67.5$^\circ$, the workspace expands, but the compliance in the central area is low, and finger interaction zone is limited. A tilt angle of 45$^\circ$ is selected as it provides the optimal balance between workspace and compliance.

\subsection{Demonstrations of the gripper}
The compliance and stiffness enable the three fingers to perform various tasks, including picking, gripping, rotating, and spinning objects of diverse shapes and sizes.

\subsubsection{Grasping}

\begin{figure}[htp]
\centering
\includegraphics[width=\linewidth]{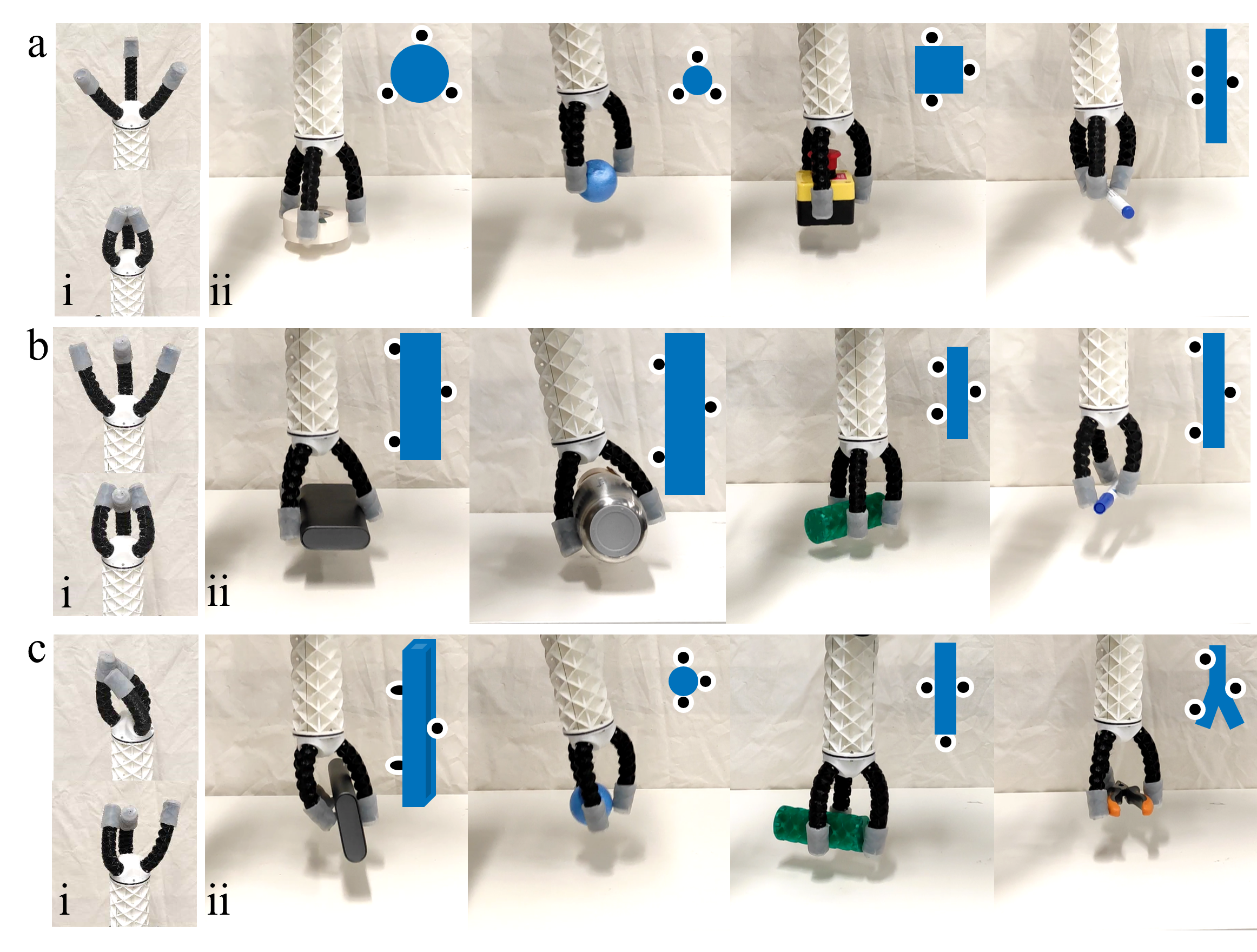}
\caption{Gripping various objects with differentpatterns}
\label{fig: Gripping}
\end{figure}

Due to the fingers' inherent flexibility, even fundamental symmetrical gripping patterns—where all fingers close at identical bending angles without lateral movement—allow the gripper to manipulate a variety of small objects (Fig. \ref{fig: Gripping}a). However, for larger objects, elongated objects, symmetrical patterns are insufficient for secure handling.
Therefore, asymmetrical configurations, wherein lateral movements are applied to the two fingers in opposite directions, are essential for enhancing the gripper’s ability to securely grasp objects along their longitudinal axes. This approach significantly increases stability, particularly when the distance between the fingers is extended (Fig. \ref{fig: Gripping}b).

Moreover, incorporating greater asymmetry into the gripper design enabled more complex tasks, such as grasping a flat object from table, clamping a spherical object between two fingers, stabilizing a cylindrical object in an alternative orientation, and grasping an irregularly shaped plier, as shown in Fig. \ref{fig: Gripping}c.

 \begin{figure}[htp]
\centering 
\includegraphics[width=\linewidth]{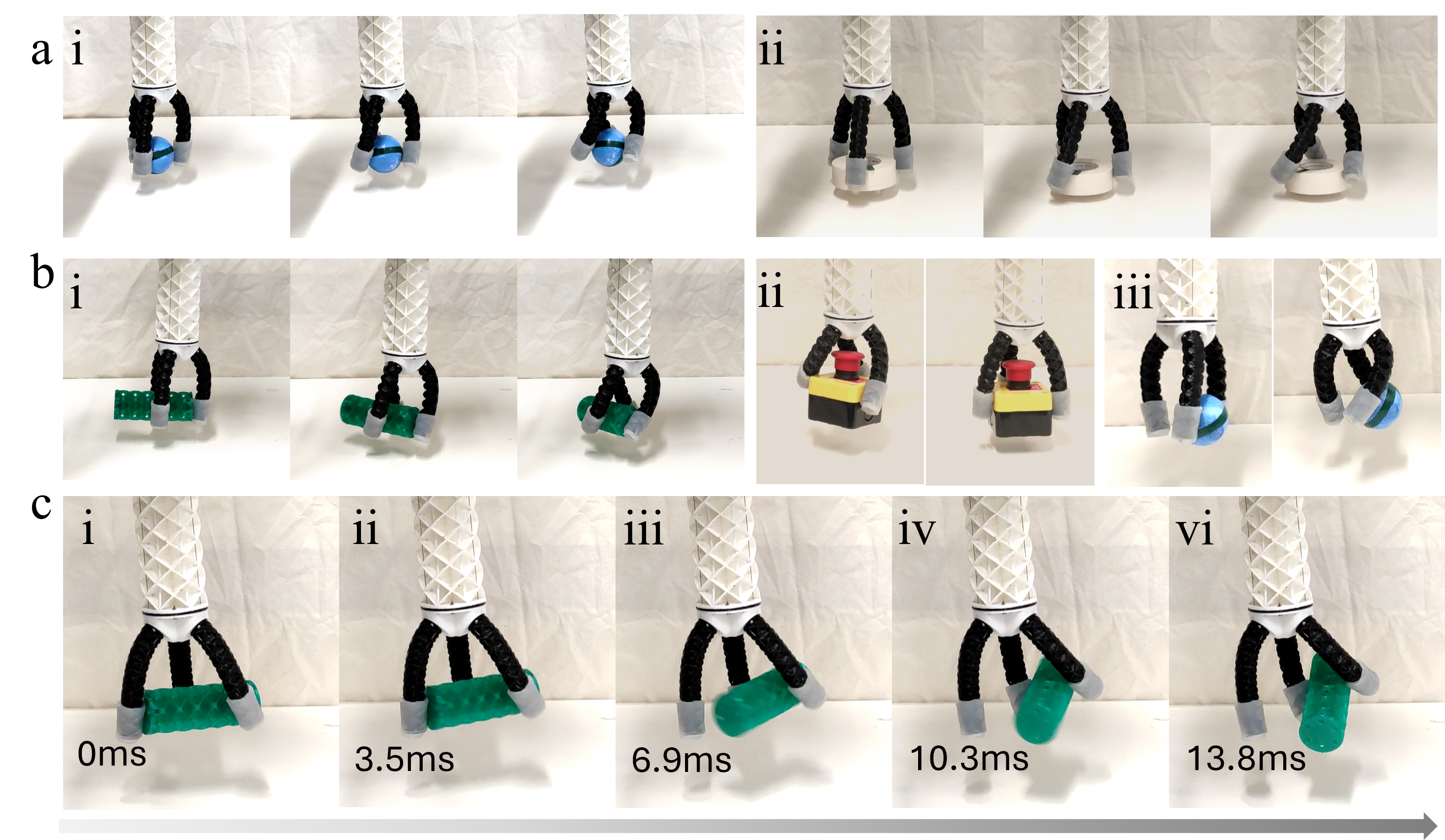}
\caption{Rotation of different shapes and gestures}
\label{fig: Rotation}
\end{figure}

\subsubsection{Rotating}
Leveraging its high degrees of freedom and compliance, the gripper can rotate round objects of varying sizes using symmetrical actuation patterns (Fig. \ref{fig: Rotation}a).  It can also gripped and rotate asymmetrical objects by combining asymmetrical gripping patterns with a twisting motion (Fig. \ref{fig: Rotation}b).

Additionally, due to the elastic energy stored in the fingers, the gripper can rotate an object within 14 ms after releasing one finger from the three-finger hold (Fig. \ref{fig: Rotation}b) showing highly dynamic behaviour.

\subsubsection{Spinning}
\begin{figure}[htp]
\centering
\includegraphics[width=\linewidth]{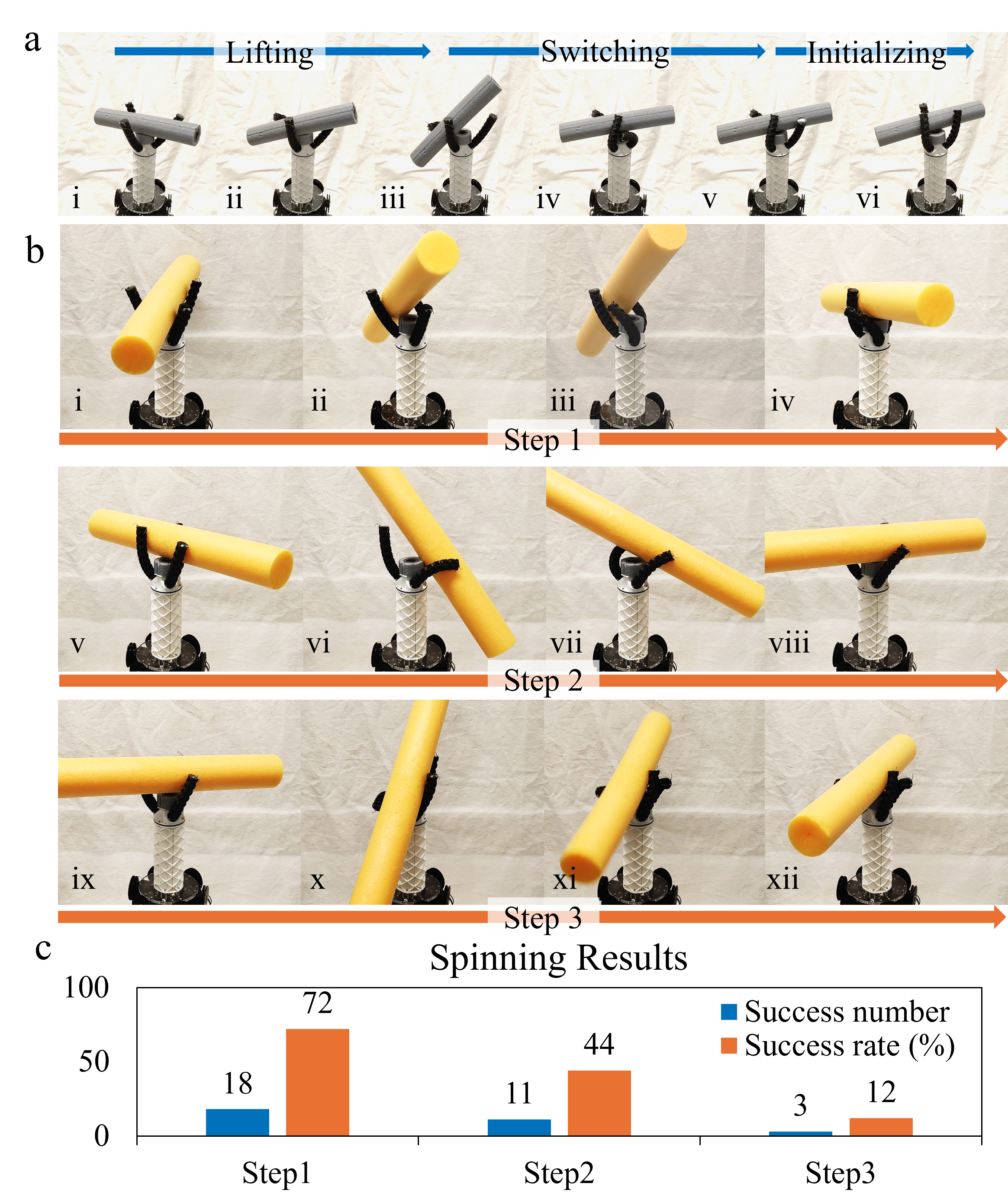}
\caption{On hand rod spinning}
\label{fig: rod spinning}
\end{figure}

As a final demonstration we combine the motion primatives to show spinning of a soft rod. 
As shown in Fig. \ref{fig: rod spinning}, the spinning process consists of three phases: lifting, switching, and initializing. To spin the foam rod, three rotational steps are required from different directions. However, the gripper's open-loop control can cause the rod’s center of gravity to shift, leading to potential failures in later steps.  The success rate for the three steps are reported in Fig. \ref{fig: rod spinning}c; out of 25 attempts, most passed the initial steps, with 3 successfully completing the full circular rotation, demonstrating the dexterity of the OTH gripper.

\section{CONCLUSIONS}
The proposed OTH structure with embedded helical tendons enables the 3D-printed TPU finger to exhibit high compliance, a large workspace, and low actuation force while storing significant energy in its deformed state. This allows the three-finger gripper to utilize tunable gripping patterns, adapting to various objects and responding swiftly to dynamic tasks. The gripper's six degrees of freedom enable complex tasks like in-hand rod spinning, which remains challenging for most advanced rigid grippers, typically requiring precise control and high degree of freedom~\cite{kumar2016optimal}.

However, the gripper is limited by open-loop control and hard-coded programming, affecting its stability and automation. Integrating vision systems and soft sensors (e.g., tactile and deformation sensors) could enhance self-perception and environmental awareness. Additionally, incorporating machine learning could further harness the gripper's dexterity. The OTH structure's rapid dynamic response also makes it suitable for use in other robots, such as legged, swimming, or snake robots.

\printbibliography

\end{document}